\documentclass{ieeeaccess}

\usepackage{cite}
\usepackage{amsmath,amssymb,amsfonts}
\usepackage{algorithmic}
\usepackage{graphicx}
\usepackage{textcomp}

\usepackage{multirow}
\usepackage{xspace}
\usepackage{balance}
\usepackage{lipsum,multicol}
\usepackage{hyperref}

\newcommand{\fig}{\@\xspace}
\newcommand{\tab}{\@\xspace}
\newcommand{\eqn}{\@\xspace}

\def\BibTeX{{\rm B\kern-.05em{\sc i\kern-.025em b}\kern-.08em
    T\kern-.1667em\lower.7ex\hbox{E}\kern-.125emX}}
\begin{document}
\history{Received December 26, 2020, accepted March 22, 2021. Date of publication xxxx 00, 0000, date of current version xxxx 00, 0000.}
\doi{10.1109/ACCESS.2021.3069041}

\title{Visual Relationship Detection with Visual-Linguistic Knowledge from Multimodal Representations}
\author{
    \uppercase{Meng-Jiun Chiou}\authorrefmark{1},
    \uppercase{Roger Zimmermann}\authorrefmark{1}, \IEEEmembership{Senior~Member,~IEEE} and 
    \uppercase{Jiashi Feng}\authorrefmark{2},~\IEEEmembership{Member,~IEEE}
}
\address[1]{Department of Computer Science, National University of Singapore. (email: \{mengjiun,rogerz\}@comp.nus.edu.sg)}
\address[2]{Department of Electrical and Computer Engineering, National University of Singapore (email: elefjia@nus.edu.sg)}
\tfootnote{This research was supported by Singapore Ministry of Education Academic Research Fund Tier 1 under MOE's official grant number T1 251RES1820.}

\markboth
{Chiou \headeretal: Visual Relationship Detection with Visual-Linguistic Knowledge from Multimodal Representations}
{Chiou \headeretal: Visual Relationship Detection with Visual-Linguistic Knowledge from Multimodal Representations}

\corresp{Corresponding authors: Meng-Jiun Chiou (e-mail: mengjiun@comp.nus.edu.sg).}

\begin{abstract}
Visual relationship detection aims to reason over relationships among salient objects in images, which has drawn increasing attention over the past few years.
Inspired by human reasoning mechanisms, it is believed that external visual commonsense knowledge is beneficial for reasoning visual relationships of objects in images, which is however rarely considered in existing methods.
In this paper, we propose a novel approach named \textit{\textbf{R}elational \textbf{V}isual-\textbf{L}inguistic \textbf{B}idirectional \textbf{E}ncoder \textbf{R}epresentations from \textbf{T}ransformers} (RVL-BERT), 
which performs relational reasoning with both visual and language commonsense knowledge learned via self-supervised pre-training with multimodal representations.
RVL-BERT also uses an effective spatial module and a novel mask attention module to explicitly capture spatial information among the objects. 
Moreover, our model decouples object detection from visual relationship recognition by taking in object names directly, enabling it to be used on top of any object detection system. 
We show through quantitative and qualitative experiments that, with the transferred knowledge and novel modules, RVL-BERT achieves competitive results on two challenging visual relationship detection datasets. 
The source code is available at \url{https://github.com/coldmanck/RVL-BERT}.
\end{abstract}


\titlepgskip=-15pt

\maketitle

\section{Introduction}

\begin{figure}[t!]
\centering
\includegraphics[width=\columnwidth]{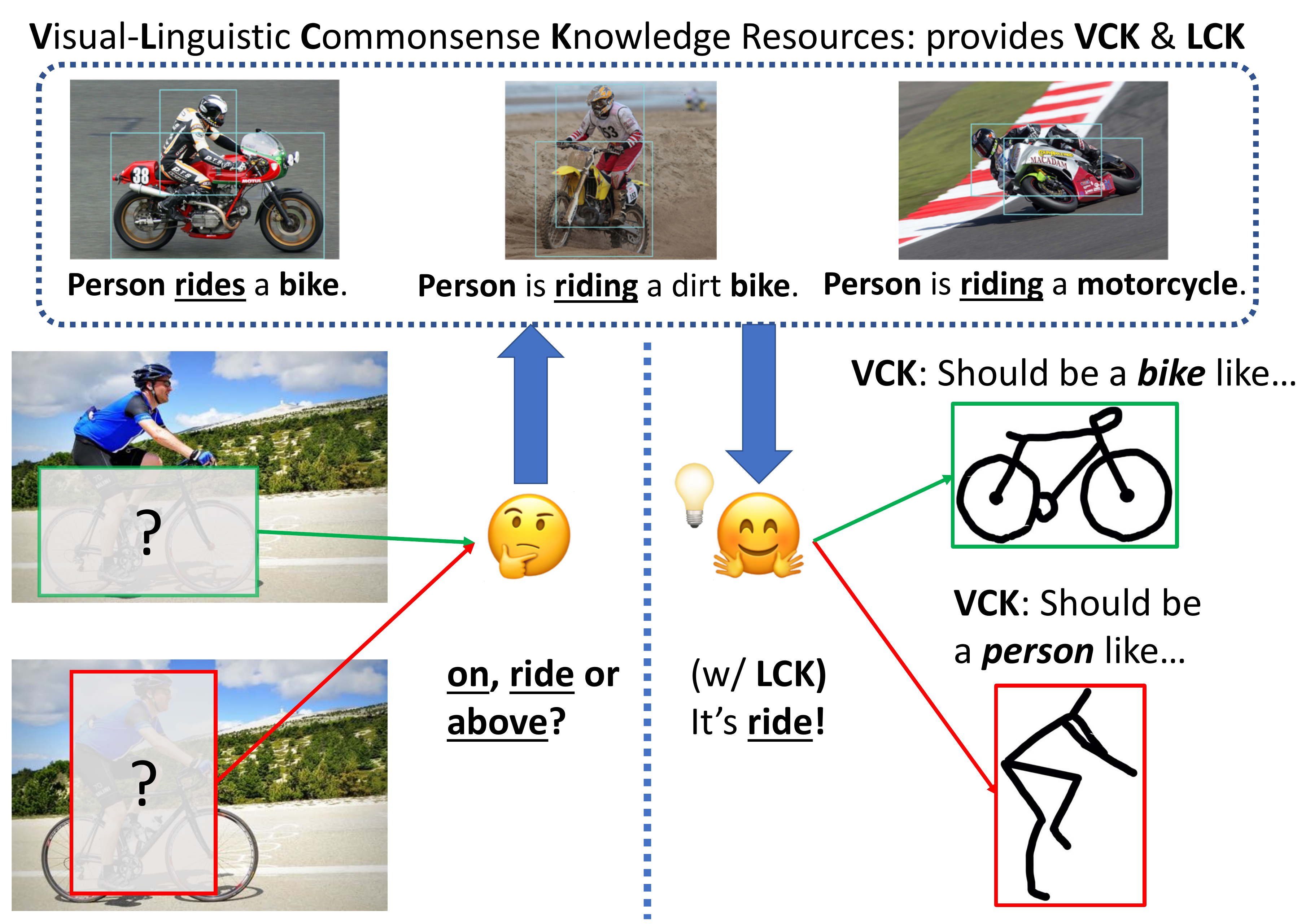}
\caption{Illustration of human reasoning over visual relationships with external visual and linguistic knowledge.
With commonsense knowledge, a human is able to ``guess'' the visually blurred regions and prefer {\fontfamily{qcr}\selectfont ride} rather than {\fontfamily{qcr}\selectfont on} or {\fontfamily{qcr}\selectfont above}.
\textbf{VCK}: Visual Commonsense Knowledge. \textbf{LCK}: Linguistic Commonsense Knowledge.}
\label{fig:intro}
\end{figure}

\IEEEPARstart{V}ISUAL relationship detection (VRD) aims to detect objects and classify triplets of {\fontfamily{qcr}\selectfont subject-predicate-object} in a query image.
It is a very crucial task for enabling an intelligent system to understand the content of images, and has received much attention over the past few years~\cite{sadeghi2011recognition,6151163,lu2016visual,li2017vip,zhuang2017towards,dai2017detecting,zhang2017visual,peyre2017weakly,yu2017visual,zhang2017ppr,yin2018zoom,jae2018tensorize,Jung2018VisualRD,hung2019union,gu2019scene,zhan2019exploring,9175559,mi2020hierarchical}.
Based on VRD, Xu et al. \cite{xu2017scene} proposed \textit{scene graph generation} (SGG) \cite{li2017scene,zellers2018neural,yang2018graph,li2018factorizable,herzig2018mapping,chen2019knowledge,zhang2019graphical,qi2019attentive,zareian2020bridging,tang2020unbiased}, which targets at extracting a comprehensive and symbolic graph representation in an image, with vertices and edges denoting instances and for visual relationships respectively.
We focus on and use the term VRD throughout this paper for consistency.
VRD is beneficial to various downstream tasks including but not limited to
image captioning \cite{li2019know,8907490},
visual question answering \cite{teney2017graph,shi2019explainable}, etc.

To enhance the performance of VRD systems, some recent works incorporate the external \textit{linguistic} commonsense knowledge from pre-trained word vectors \cite{lu2016visual}, structured knowledge bases~\cite{gu2019scene}, raw language corpora~\cite{yu2017visual}, etc.,
as priors,
which has taken inspiration from human reasoning mechanism. 
For instance, for a relationship triplet case {\fontfamily{qcr}\selectfont person-ride-bike} as shown in Figure\fig\ref{fig:intro}, with linguistic commonsense, the predicate {\fontfamily{qcr}\selectfont ride} is more accurate for describing the relationship of {\fontfamily{qcr}\selectfont person} and {\fontfamily{qcr}\selectfont bike}
compared with other relational descriptions like {\fontfamily{qcr}\selectfont on} or {\fontfamily{qcr}\selectfont above}, which are rather abstract. 
In addition, we argue that the external \textit{visual} commonsense knowledge is also beneficial to lifting detection performance of the VRD models, which is however rarely considered previously.
Take the same {\fontfamily{qcr}\selectfont person-ride-bike} in Figure\fig\ref{fig:intro} as an example.
If the pixels inside the bounding box of {\fontfamily{qcr}\selectfont person} are masked (zeroed) out, humans can still predict them as a person since we have seen many examples and have plenty of visual commonsense regarding such cases.
This reasoning process would be helpful for VRD systems since it incorporates relationships of the basic visual elements; however, most previous approaches learn visual knowledge only from target datasets and neglect external visual commonsense knowledge in abundant unlabeled data. 
Inspired by the recent successful visual-linguistic pre-training methods (BERT-like models) \cite{li2019visualbert,lu2019vilbert}, 
we propose to exploit both \textit{linguistic} and \textit{visual} commonsense knowledge from Conceptual Captions \cite{sharma2018conceptual} --- a large-scale dataset containing 3.3M images with coarsely-annotated descriptions (alt-text) that were crawled from the web, to achieve boosted VRD performance. 
We first pre-train our backbone model (multimodal BERT) on Conceptual Captions with different pretext tasks to learn the visual and linguistic commonsense knowledge. 
Specifically, our model mines visual prior information via learning to predict labels for an image's subregions that are randomly masked out.
The model also considers linguistic commonsense knowledge through learning to predict randomly masked out words of sentences in image captions.
The pre-trained weights are then used to initialize the backbone model and trained together with other additional modules (detailed at below) on visual relationship datasets.

Besides visual and linguistic knowledge, spatial features are also important cues for reasoning over object relationships in images. 
For instance, for {\fontfamily{qcr}\selectfont A}-{\fontfamily{qcr}\selectfont on}-{\fontfamily{qcr}\selectfont B}, the bounding box (or it's center point) of {\fontfamily{qcr}\selectfont A} is often above that of {\fontfamily{qcr}\selectfont B}. 
However, such spatial information is not explicitly considered in BERT-like visual-linguistic models \cite{su2019vl,lu2019vilbert,li2019visualbert}. 
We thus design two additional modules to help our model better utilize such information: a mask attention module and a spatial Module. 
The former predicts soft attention maps of target objects, which are then used to enhance visual features by focusing on target regions while suppressing unrelated areas;
the latter augments the final features with bounding boxes coordinates to explicitly take spatial information into account.

We integrate the aforementioned designs into a novel VRD model, 
named \textit{\textbf{R}elational \textbf{V}isual-\textbf{L}inguistic \textbf{B}idirectional \textbf{E}ncoder \textbf{R}epresentations from \textbf{T}ransformers} (RVL-BERT). 
RVL-BERT makes use of the pre-trained visual-linguistic representations as the source of visual and language knowledge to facilitate the learning and reasoning process on the downstream VRD task. 
It also incorporates a novel mask attention module to actively focus on the object locations in the input images and a spatial module to capture spatial relationships more accurately.
Moreover, RVL-BERT is flexibxle in that it can be placed on top of any object detection model. 

Our contribution in this paper is three-fold.
Firstly, we are among the first to identify the benefit of visual-linguistic commonsense knowledge to visual relationship detection, especially when objects are occluded.
Secondly, we propose RVL-BERT -- a multimodal VRD model pre-trained on visaul-linguistic commonsense knowledge bases learns to predict visual relationships with the attentions among visual and linguistic elements, with the aid of the spatial and mask attention module.
Finally, we show through extensive experiments that the commonsense knowledge and the proposed modules effectively improve the model performance, and our RVL-BERT achieves competitive results on two VRD datasets.

\section{Related Work}
\label{sec:related-work}

\subsection{Visual Relationship Detection}
Visual relationship detection (VRD) is a task reasoning over the relationships between salient objects in the images. 
Understanding relationships between objects is essential for other vision tasks such as action detection \cite{8848601,7589110} and image retrieval \cite{johnson2015image,schuster2015generating}.
Recently, \textit{linguistic knowledge} has been incorporated as guidance signals for the VRD systems.
For instance, \cite{lu2016visual} proposed to detect objects and predicates individually with language priors and fuse them into a higher-level representation for classification. 
\cite{dai2017detecting} exploited statistical dependency between object categories and predicates to infer their subtle relationships.
Going one step further, \cite{gu2019scene} proposed a dedicated module utilizing bi-directional Gated Recurrent Unit to encode \textit{external} language knowledge and a Dynamic Memory Network \cite{kumar2016ask} to pick out the most relevant facts. 

However, none of the aforementioned works consider external \textit{visual} commonsense knowledge, which is also beneficial to relationship recognition.
By contrast, we propose to exploit the abundant visual commonsense knowledge from multimodal Transformers \cite{vaswani2017attention} learned in pre-training tasks to facilitate the relationship detection in addition to the linguistic prior.

\subsection{Representation Pre-training}
In the past few years, self-supervised learning which utilizes its own unlabeled data for supervision has been widely applied in representation pre-training. 
BERT, ELMo \cite{peters2018deep} and GPT-3 \cite{brown2020language} are representative language models that perform self-supervised pre-training on various pretext tasks with either Transformer blocks or bidirectional LSTM.
More recently, increasing attention has been drawn to multimodal (especially visual and linguistic) pre-training. 
Based on BERT, Visual-Linguistic BERT (VL-BERT) \cite{su2019vl} pre-trains a single stream of cross-modality transformer layers from not only image captioning datasets but also language corpora.
It is trained on BooksCorpus \cite{zhu2015aligning} and English Wikipedia in addition to Conceptual Captions \cite{sharma2018conceptual}.
We refer interested readers to \cite{su2019vl} for more details of VL-BERT.

In this work, we utilize both visual and linguistic commonsense knowledge learned in the pretext tasks. 
While VL-BERT can be applied to training VRD without much modification, we show experimentally that their model does not perform well due to lack of attention to spatial features. 
By contrast, we propose to enable knowledge transfer for boosting detection accuracy and use two novel modules to explicitly exploit spatial features. 

\section{Methodology}
\label{sec:methodology}
\subsection{Revisiting BERT and VL-BERT}
Let a sequence of $N$ embeddings $x = \{x_1, x_2, ..., x_N\}$ be the features of input sentence words, which are the summation of \textit{token}, \textit{segment} and \textit{position embedding} as defined in BERT \cite{devlin2018bert}. 
The BERT model takes in $x$ and utilizes a sequence of $n$ multi-layer bidirectional Transformers \cite{vaswani2017attention} to learn contextual relations between words. 
Let the input feature at layer $l$ denoted as $x^l = \{x^l_1, x^l_2, ..., x^l_N\}$. 
The feature of x at layer $(l+1)$, denoted as $x^{l+1}$, is computed through a Transformer layer which consists of two sub-layers: 
1) a multi-head self-attention layer plus a residual connection
\begin{align}
\tilde{h}^{l+1}_i &= \sum_{m=1}^{M} W^{l+1}_m  \Big\{ \sum_{j=1}^{N} A^{m}_{i,j} \cdot V_m^{l+1} x_j^l \Big\}, \label{eq:multi_head} \\
h^{l+1}_i &= \text{LayerNorm} (x^l_i + \tilde{h}^{l+1}_i), \label{eq:ffn1}
\end{align}
where $A_{i,j}^m \propto (Q_m^{l+1}x_i^l)^T (K_m^{l+1}x_j^l)$ represents a normalized dot product attention mechanism between the $i$-th and the $j$-th feature at the $m$-th head, 
and 2) a position-wise fully connected network plus a residual connection
\begin{align}
\tilde{x}^{l+1}_i &= W_2^{l+1}\cdot\text{GELU} ( (W_1^{l+1} h^{l+1}_i) + b_1^{l+1} ) + b_2^{l+1}, \label{eq:feedfoward} \\
x^{l+1}_i &= \text{LayerNorm} (h^{l+1}_i + \tilde{x}^{l+1}_i), \label{eq:ffn2}
\end{align}
where GELU is an activation function named Gaussian Error Linear Unit \cite{hendrycks2016gaussian}. 
Note that $\mathbf{Q}$ (Query), $\mathbf{K}$ (Key), $\mathbf{V}$ (Value) are learnable embeddings for the attention mechanism, and $\mathbf{W}$ and $\mathbf{b}$ are learnable weights and biases respectively.

Based on BERT, VL-BERT \cite{su2019vl} adds $O$ more multi-layer Transformers to take in additional $k$ visual features. 
The input embedding becomes $x = \{x_1, ..., x_N, x_{N+1},..., x_{N+O}\}$, which is computed by the summation of not only the token, segment and position embeddings but also an additional \textit{visual feature embedding} which is generated from the bounding box of each corresponding word. 
The model is then pre-trained on two types of pretext tasks to learn the visual-linguistic knowledge: 1) \textit{masked language modeling with visual clues} that predicts a randomly masked word in a sentence with image features, and 2) \textit{masked RoI classification with linguistic clues} that predicts the category of a randomly masked region of interest (RoI) with linguistic information.

\subsection{Overview of Proposed Model}

\begin{figure*}[t!]
\centering
\includegraphics[width=\textwidth]{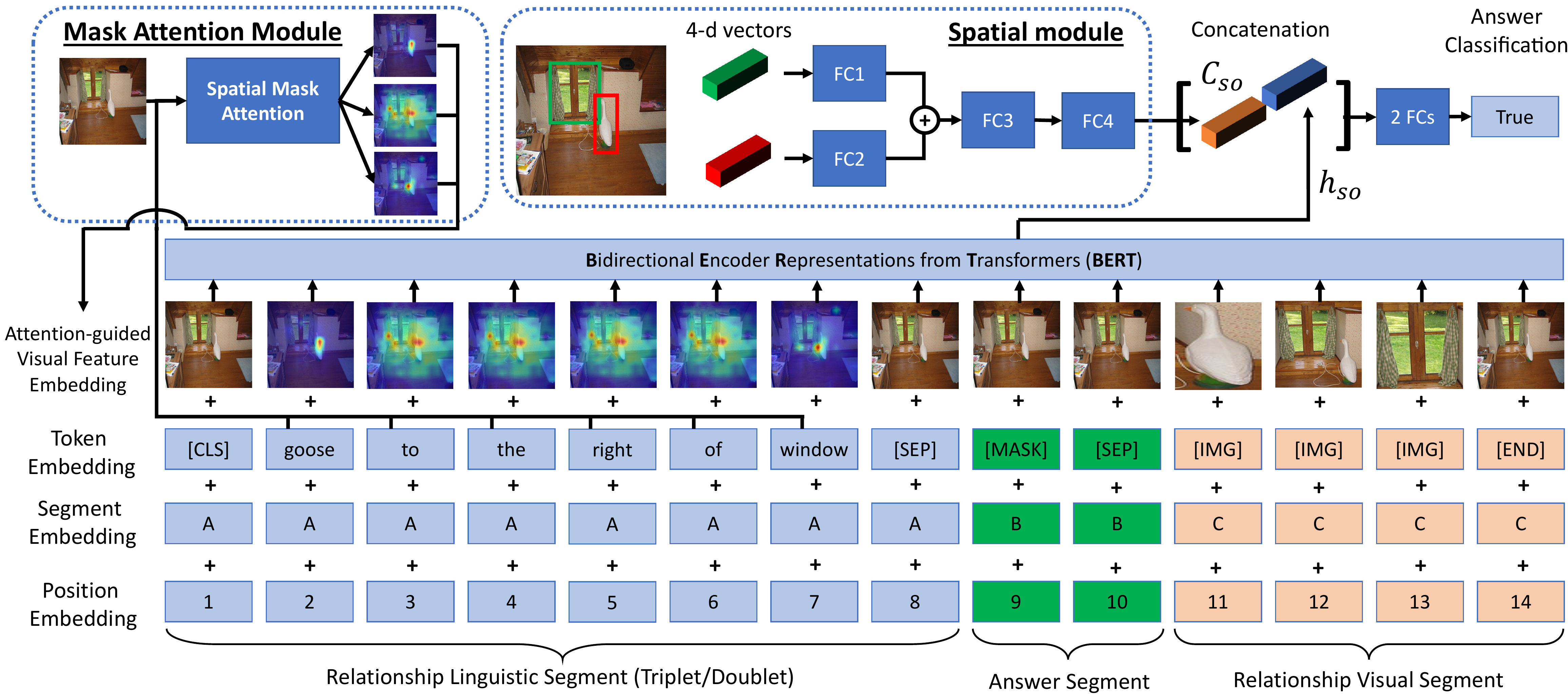}
\caption{Architecture illustration of proposed RVL-BERT for SpatialSense dataset \cite{yang2019spatialsense}. 
It can be easily adapted for VRD dataset \cite{lu2016visual} by replacing triplets {\fontfamily{qcr}\selectfont subject-predicate-object} with doublets {\fontfamily{qcr}\selectfont subject-object} and performing predicate classification instead of binary classification on the output feature of ``[MASK]''.}
\label{fig:rvl-bert}
\end{figure*}

Figure\fig\ref{fig:rvl-bert} shows the overall architecture of our proposed RVL-BERT. 
For the backbone BERT model, we adopt a 12-layer Transformer and initialize it with the pre-trained weights of VL-BERT for visual and linguistic commonsense knowledge.
Note that while our model is inspired by VL-BERT, it differs in several important aspects:
1) RVL-BERT explicitly arranges query object pairs in sequences of {\fontfamily{qcr}\selectfont subject-predicate-object} (instead of sentences in the original design) and receives an extra answer segment for relationship prediction.
2) Our model is equipped with a novel mask attention module that learns attention-guided visual feature embeddings for the model to attend to target object-related area.
3) A simple yet effective spatial module is added to capture spatial representation of subjects and objects, which are of importance in spatial relationship detection. 

Let $N$, $A$ and $O$ denote the number of elements for the relationship linguistic segment, the answer segment, and the relationship visual segment, respectively.
Our model consists of $N + A + O$ multi-layer Transformers, which takes in a sequence of linguistic and visual elements, including the output from the mask attention module, and learns the context of each element from all of its surrounding elements. 
For instance, as shown in Figure\fig\ref{fig:rvl-bert}, to learn the representation of the linguistic element {\fontfamily{qcr}\selectfont goose}, 
the model looks at not only the other linguistic elements (\emph{e.g.}, {\fontfamily{qcr}\selectfont to the right of} and {\fontfamily{qcr}\selectfont window}) but also all visual elements (\emph{e.g.}, {\fontfamily{qcr}\selectfont goose}, {\fontfamily{qcr}\selectfont window}). 
Along with the multi-layer Transformers, the spatial module extracts the location information of subjects and objects using their bounding box coordinates.
Finally, the output representation of the element in the answer segment, $h_{so}$, is augmented with the output of the spatial module $C_{so}$, followed by classification with a 2-layer fully connected network.

The input to the model can be divided into three groups by the type of segment, or four groups by the type of embedding. We explain our model below from the segment-view and the embedding-view, respectively.

\subsubsection{Input Segments}
For each input example, RVL-BERT receives a relationship linguistic segment, an answer segment, and a relationship visual segment as input.
\\\\
\noindent a) \textit{Relationship linguistic segment} (light blue elements in Figure\fig\ref{fig:rvl-bert}) is the linguistic information in a triplet form {\fontfamily{qcr}\selectfont subject-predicate-object}, like the input form of SpatialSense dataset \cite{yang2019spatialsense}, or a doublet form {\fontfamily{qcr}\selectfont subject-object} like the input in VRD dataset \cite{lu2016visual}). 
Note that each term in the triplet or doublet may have more than one element, such as {\fontfamily{qcr}\selectfont to the right of}. This segment starts with a special element ``[CLS]'' that stands for classification\footnote{We follow the original VL-BERT to start a sentence with the ``[CLS]'' token, but we do not use it for classification purposes.} and ends with a ``[SEP]'' that keeps different segments separated.
\\\\
\noindent b) \textit{Answer segment} (green elements in Figure\fig\ref{fig:rvl-bert}) is designed for learning a representation of the whole input and has only special elements like ``[MASK]'' that is for visual relationship prediction and the same ``[SEP]'' as in the relationship linguistic segment.
\\\\
\noindent c) \textit{Relationship visual segment} (tangerine color elements in Figure\fig\ref{fig:rvl-bert}) is the visual information of a relationship instance, also taking the form of triplets or doublets but with each component term corresponding to only one element even if its number of words of the corresponding label is greater than one. 

\subsubsection{Input Embeddings}
There are four types of input embeddings: token embedding $t$, segment embedding $s$, position embedding $p$, and (attention-guided) visual feature embedding $v$.
Among them, the attention-guided visual feature embedding is newly introduced while the others follow the original design of VL-BERT.
We denote the input of RVL-BERT as $x = \{x_1, ..., x_N,$ $x_{N+1}, ..., x_{N+A},$ $x_{N+A+1}...,x_{N+A+O}\}$, $\forall x_i:$ $x_i=t_i+v_i+s_i+p_i$ where $t_i \in t$, $v_i \in v$, $s_i \in s$, $p_i \in p$.
\\\\
\noindent a) \textit{Token Embedding}.
We transform each of the input words into a $d$-dimensional feature vector using WordPiece embeddings \cite{wu2016google} comprising $30,000$ distinct words. 
In this sense, our model is flexible since it can take in any object label with any combination of words available in WordPiece. 
Note that for those object/predicate names with more than one word, the exact same number of embeddings is used. 
For the $i$-th object/predicate name in an input image, we denote the token embedding as $t = \{t_1, ..., t_N, t_{N+1}, ...,$ $ t_{N+A}, t_{N+A+1}..., t_{N+A+O}\}$, $t_i \in \mathbb{R}^{d}$, where $d$ is the dimension of the embedding. 
We utilize WordPiece embeddings for relationship triplets/doublets $\{t_2,...,t_{N-1}\}$, and use special predefined tokens ``[CLS]'', ``[SEP]'', ``[MASK]'' and ``[IMG]'' for the other elements.
\\\\
\noindent b) \textit{Segment Embedding}.
We use three types of learnable segment embeddings $s=\{s_1,...,s_{N+A+O}\}, s_i \in \mathbb{R}^d$ to inform the model that there are three different segments: "$A$" for relationship linguistic segment, "$B$" for answer segment and "$C$" for relationship visual segment. 
\\\\
\noindent c) \textit{Position Embedding}.
Similar to segment embeddings, learnable position embeddings $p=\{p_1,...,$ $p_{N+A+O}\}, p_i \in \mathbb{R}^d$ are used to indicate the order of elements in the input sequence. 
Compared to the original VL-BERT where the position embeddings of the relationship visual segment are the same for each RoI, we use distinct embeddings as our RoIs are distinct and ordered. 
\\\\
\noindent d) \textit{Visual Feature Embedding}. These embeddings are to inform the model of the internal visual knowledge of each input word.
Given an input image and a set of RoIs, a CNN backbone is utilized to extract the feature map, which is prior to the output layer, followed by RoI Align \cite{he2017mask} to produce fixed-size feature vectors $z=\{z_0, z_1,...,z_K\}, z_i \in \mathbb{R}^d$ for $K$ RoIs, where $z_0$ denotes the feature of the whole image. 
For triplet inputs, we additionally generate $K(K-1)$ features for all possible union bounding boxes: $u = \{u_1,...,u_{K(K-1)}\}, u_i \in \mathbb{R}^d$. 
We denote the input visual feature embedding as $v = \{v_1, ..., v_N,$ $v_{N+1}, ..., v_{N+A}, v_{N+A+1}..., v_{N+A+O}\}$, $v_i \in \mathbb{R}^{d}$. 
We let subject and object be $s$ and $o$, with $s, o \in \{1,...,K\}, s \neq o$, and let the union bounding box of $s$ and $o$ be $so \in \{1,...,K(K-1)\}$.

For the relationship visual segment $\{v_{N+A},...,$ $v_{N+A+O-1}\}$ (excluding the final special element), we use $z_s$ and $z_o$ as the features of subject $s$ and object $o$ in doublet inputs, and add another $u_{so}$ in between in case of triplet inputs.
For the special elements other than ``[IMG]'', we follow VL-BERT to use the full image feature $z_0$. 
However, for the relationship linguistic segment $\{v_{2},...,v_{N-1}\}$ (excluding the first and final special elements), it is unreasonable to follow the original design to use the same, whole-image visual feature for all elements, since each object/predicate name in the relationship linguistic segment should correspond to different parts of the image.
To better capture distinct visual information for elements in relationship linguistic segment, we propose a \textit{mask attention module} to learn to generate attention-guided visual feature embeddings that attend to important (related) regions, which is detailed at below.

\subsection{Mask Attention Module}

\begin{figure}[t!]
\centering
\includegraphics[width=\columnwidth]{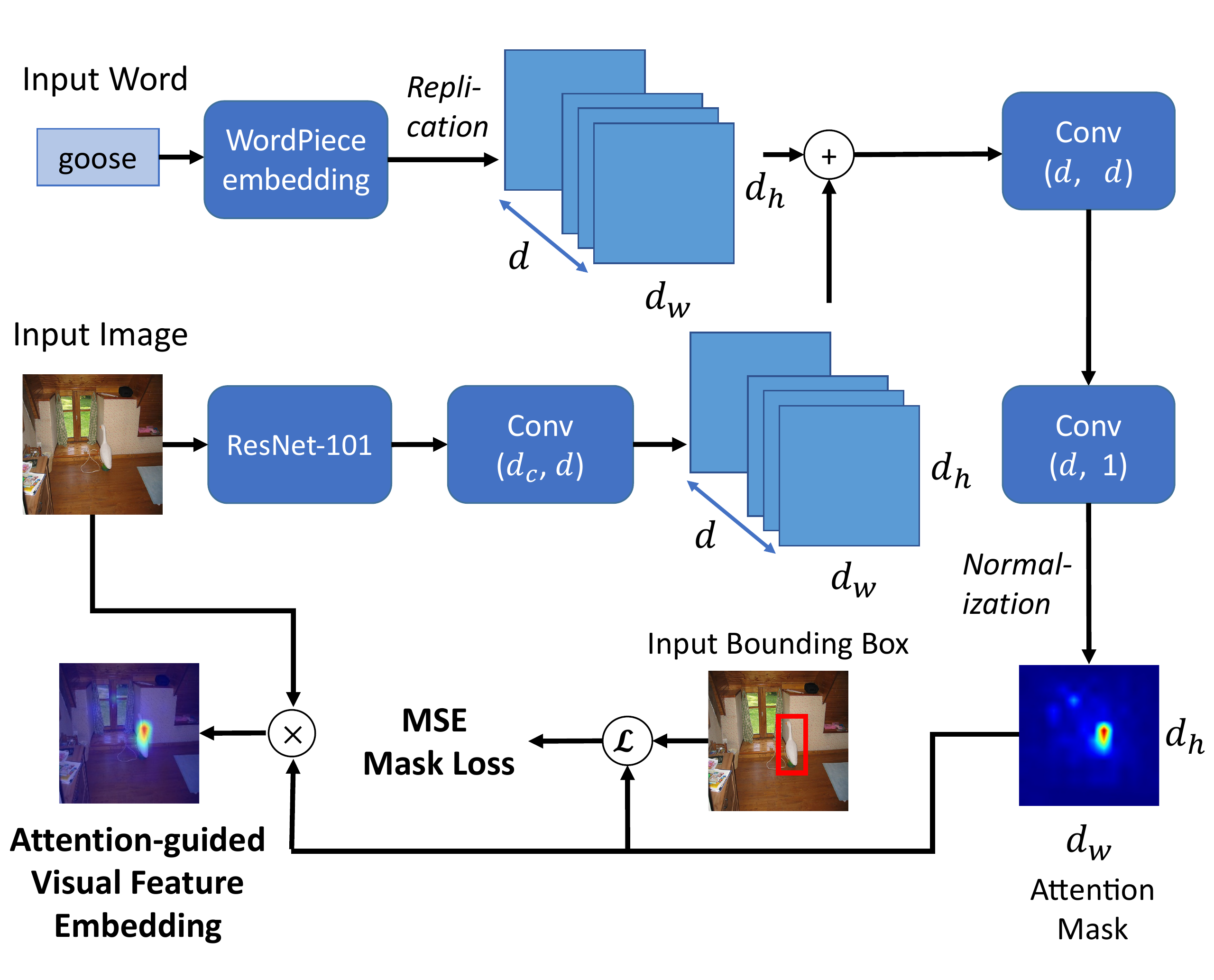}
\caption{The pipeline of Mask Attention Module. Given an input image and the corresponding word embedding(s), the module generates an attention mask (heatmap) and outputs an attention-guided visual embedding.}
\label{fig:mask-attention-module}
\end{figure}

An illustration of the mask attention module is shown in Figure\fig\ref{fig:mask-attention-module}. 
Denote the visual feature (the feature map before average pooling) used by the mask attention module as $v_s \in \mathbb{R}^{d_c \times d_w \times d_h}$, where $d_c, d_w, d_h$ stand for the dimension of the channel, width, and height, respectively. 
To generate the feature for an object $s$ (\emph{e.g.}, {\fontfamily{qcr}\selectfont goose} in Figure\fig\ref{fig:mask-attention-module}),
the mask attention module takes in and projects the visual feature $v_s$ and the word embedding
\footnote{Note that for object labels with more than one word, the embeddings of each word are element-wise summed in advance.}
$w_s$ into the same dimension using a standard CNN and a replication process, respectively
\begin{align}
    \tilde{v}_s &= \sigma( W_1^T v_s + b_1), \\
    w_s &= \text{Replication}(w_s),
\end{align}
where $\text{Replication}(\cdot)$ replicates the input vector of size $d$ into the feature map of dimension $d \times d_w \times d_h$. 
The above is followed by element-wise addition to fuse the features, two convolutional layers as well as a re-scaling process to generate the attention mask $m_s$
\begin{align}
    \tilde{m_s} &= \sigma(W_2^T (\tilde{v}_s + w_s) + b_2), \\
    m_s &= \text{Norm}(W_3^T \tilde{m_s} + m_3),
\end{align}
where the min-max $\text{Norm}(\cdot)$ applied to each element is defined by $\text{Norm}(x_i) = \frac{x_i - \min(x)}{\max(x) - \min(x)}$.
Note that in the above equations all of the $W$'s and $b$'s are learnable weights and biases of the convolutional layers, respectively. 
The attention-guided visual feature $v_s^{att}$ is then obtained by performing Hadamard product between the visual feature and the attention mask: $v_s^{att} = v_s \circ m_s$. 
Finally, $v_s^{att}$ is pooled into $v_s^{att} \in \mathbb{R}^d$ to be used in $\{v_2,...,v_{N-1}\}$.

To learn to predict the attention masks, we train the module against the Mean Squared Error (MSE loss) between the mask $m_s$ and the resized ground truth mask $b_s$  consisting of all ones inside the bounding box and outside all zeros:
\begin{equation}
    \mathcal{L}_{mask} = \frac{1}{d_wd_h} \sum_{i=1}^{d_w} \sum_{j=1}^{d_h} (m_s^{ij} - b_s^{ij})^2,
\end{equation}
where $d_w$, $d_h$ denote the width and length of the attention mask. 

\subsection{Spatial Module}
The spatial module aims to augment the output representation with spatial knowledge by paying attention to bounding box coordinates. 
See the top part of Figure\fig\ref{fig:rvl-bert} for its pipeline.

Let $(x_i^0, y_i^0)$, $(x_i^1, y_i^1)$ denote the top-left and bottom-right coordinates of a bounding box of an object $i$ of an input image, and let $w$, $h$ be the width and height of the image.
The 4-dimensional normalized coordinate of an object $i$ is defined by $C_i = (x_i^0/w, y_i^0/h, x_i^1/w, y_i^1/h)$. 
The spatial module takes in coordinate vectors of a subject $s$ and an object $o$, and encodes them using linear layers followed by element-wise addition fusion and a two-layer, fully-connected layer 
\begin{align}
\tilde{C}_{so} &= \sigma(W_{4} C_s + b_{4}) + \sigma(W_{5} C_o + b_{5}), \\
C_{so} &= W_{7}\ \sigma(W_{6} \tilde{C}_{so} + b_{6}) + b_7.
\end{align}
The output feature $C_{so}$ is then concatenated with the multimodal feature $h_{so}$ to produce $f_{so}$ for answer classification:
\begin{equation}
    f_{so} = [C_{so}; h_{so}].
    \label{eq:feature_combination}
\end{equation}

\section{Experiments}

\subsection{Datasets}
\label{sec:datasets}
We first ablate our proposed model on VRD dataset \cite{lu2016visual}, which is the most widely used benchmark.
For comparison with previous methods, we also evaluate on SpatialSense \cite{yang2019spatialsense} dataset. 
Compared with Visual Genome (VG) dataset \cite{krishna2017visual}, SpatialSense suffers less from the dataset language bias problem, which is considered a distractor for performance evaluation --- in VG, the visual relationship can be ``guessed'' even without looking at the input image \cite{zellers2018neural,chen2019knowledge}.\footnote{Despite the strong dataset bias, we still provide the experimental results on VG in appendix for reference.}

\def\arraystretch{1.2}

\subsubsection{VRD}
The VRD dataset consists of 5,000 images with 37,993 visual relationships. 
We follow \cite{lu2016visual} to divide the dataset into a training set of 4,000 images and a test set of 1,000 images, while only 3,780 and 955 images are annotated with relations, respectively. 
For all possible pairs of objects in an image, our model predicts by choosing the best-scoring predicate and records the scores, which are then used to rank all predictions in the ascending order. 
Since the visual relationship annotations in this dataset are far from exhaustive, we cannot use precision or average precision as they will penalize correct detections without corresponding ground truth. 
Traditionally, Recall@K is adopt to bypass this problem and we follow this practice throughout our experiments.
For VRD, the task named \textit{Predicate Detection/Classification} measures the accuracy of predicate prediction given ground truth classes and bounding boxes of subjects and objects independent of the object detection accuracy.
Following \cite{lu2016visual,zhang2017visual}, we use \textbf{Recall@}$\mathbf{K}$, or the fraction of ground truth relations that are recalled in the top $K$ candidates. 
$K$ is usually set as 50 or 100 in the literature.

\subsubsection{SpatialSense}
SpatialSense is a relatively new visual relationship dataset focusing on especially spatial relations.
Different from Visual Genome \cite{krishna2017visual}, SpatialSense is dedicated to reducing dataset bias, via a novel data annotation approach called Adversarial Crowdsourcing which prompts annotators to choose relation instances that are hard to guess by only looking at object names and bounding box coordinates. 
SpatialSense defines nine spatial relationships {\fontfamily{qcr}\selectfont above}, 
{\fontfamily{qcr}\selectfont behind}, 
{\fontfamily{qcr}\selectfont in}, 
{\fontfamily{qcr}\selectfont in front of}, 
{\fontfamily{qcr}\selectfont next to},
{\fontfamily{qcr}\selectfont on},
{\fontfamily{qcr}\selectfont to the left of},
{\fontfamily{qcr}\selectfont to the right of}, and
{\fontfamily{qcr}\selectfont under}, and contains 17,498 visual relationships in 11,569 images. 
The task on SpatialSense is binary classification on given visual relationship triplets of images, namely judging if a triplet {\fontfamily{qcr}\selectfont subject-predicate-object} holds for the input image.
Since in SpatialSense the number of examples of ``True'' equals that of ``False'', the \textbf{classification accuracy} can be used as a fair measure.
We follow the original split in \cite{yang2019spatialsense} to divide them into 13,876 and 3,622 relations for training and test purposes, respectively. 

\subsection{Implementation}
For the backbone model, we use BERT\textsubscript{BASE}\footnote{There are two variants of BERT: BERT\textsubscript{BASE} that has 12-layer Transformers and BERT\textsubscript{LARGE} that has 24-layer Transformers.} that is pre-trained on three datasets including Conceptual Captions \cite{sharma2018conceptual}, BooksCorpus \cite{zhu2015aligning} and English Wikipedia.
For extracting visual embedding features, we adopt Fast R-CNN \cite{girshick2015fast} (detection branch of Faster R-CNN \cite{ren2015faster}).
We randomly initialize the final two fully connected layers and the newly proposed modules (\emph{i.e.}, mask attention module and spatial module).
During training, we find our model empirically gives the best performance when freezing the parameters of the backbone model and training on the newly introduced modules.
We thus get a lightweight model compared to the original VL-BERT as the number of trainable parameters is reduced by around 96\%, \emph{i.e.}, down from 161.5M to 6.9M and from 160.9M to 6.4M when trained on the SpatialSense dataset and the VRD dataset, respectively. 
ReLU is used as the nonlinear activation function $\sigma$. 
We use $d = 768$ for all types of input embeddings, $d_c = 2048$ for the dimension of channel of the input feature map and $d_w = d_h = 14$ for the attention mask in the mask attention module. 
The training loss is the sum of the softmax cross-entropy loss for answer classification and the MSE loss for the mask attention module.
The experiments were conducted on a single NVIDIA Quadro RTX 6000 GPU in an end-to-end manner using Adam \cite{kingma2014adam} optimizer with initial learning rate $1 \times 10^{-4}$ after linear warm-up over the first $500$ steps, weight decay $1 \times 10^{-4}$ and exponential decay rate 0.9 and 0.999 for the first- and the second-moment estimates, respectively. 
We trained our model for 60 and 45 epochs for VRD and SpatialSense dataset, respectively, as there are more images in the training split of SpatialSense.
For experiments on the VRD dataset, we followed the training practice in \cite{hung2019union} to train with an additional ``no relationship'' predicate and for each image we sample $32$ relationships with the ratio of ground truth relations to negative relations being $1:3$.

\subsection{Ablation Study Results}

\subsubsection{Training Objective for Mask Attention Module}
We first compare performance difference between training the mask attention module (MAM) against \textbf{MSE} loss or binary cross entropy (\textbf{BCE}) loss. 
The first two rows of Table\tab\ref{table:ablation_studies} show that MSE outperforms BCE by relative 3.8\% on Recall@50. 
We also observe that training with BCE is relatively unstable as it is prone to gradient explosion under the same setting.

\begin{table}[t!]
    \def\arraystretch{1.2}
    \centering
    \caption{Ablation results for different losses of mask attention module and ways of feature combination. \textbf{.3}, \textbf{.5} and \textbf{.7} denote different $\alpha$ values in $f_{so} = \alpha C_{so} + (1 - \alpha) h_{so}$.}
    \label{table:ablation_studies}
    \resizebox{0.9\linewidth}{!}{$
    \begin{tabular}{cccccc|c}
    \hline
    \multicolumn{2}{c}{MAM Loss} & \multicolumn{4}{c|}{Feature Combination} & \multirow{2}{*}{Recall@50} \\
    BCE & MSE & .3 & .5 & .7 & concat & \\
    \hline
    \checkmark & & & & & \checkmark & 53.50 \\
    & \checkmark & & & & \checkmark & \textbf{55.55} \\
    & \checkmark & \checkmark & & & & 55.46 \\
    & \checkmark & & \checkmark & & & 54.74 \\
    & \checkmark & & & \checkmark & & 55.19 \\
    \hline
    \end{tabular}
    $}
\end{table}

\subsubsection{Feature Combination}

We also experiment with different ways of feature combination, namely, element-wise addition and concatenation of the features. 
To perform the experiments, we modify Eqn.\eqn\ref{eq:feature_combination} as $f_{so} = \alpha C_{so} + (1 - \alpha) h_{so}$,
and we experiment with different $\alpha$ values (\textbf{.3}, \textbf{.5} and \textbf{.7}). 
The last five rows of Table\tab\ref{table:ablation_studies} show that concatenation performs slightly better than addition under all $\alpha$ values.

\begin{table}[t!]
    \begin{center}
    \caption{Ablation results of each module on VRD dataset (Recall@50) and SpatialSense dataset (Overall Acc.) \textbf{VL}: Visual-Linguistic Knowledge. \textbf{S}: Spatial. \textbf{M}: Mask Att.}
    \label{table:module_effeciveness}
    \def\arraystretch{1.2}
    \resizebox{1.0\linewidth}{!}{$
    \begin{tabular}{l|ccc|c|c}
    \hline
    Model & VL & Spatial & Mask Att. & R@50 & Acc. \\
    \hline
    Basic & & & & 40.22 & 55.4 \\
    +VL & \checkmark & & & 45.06 & 61.8 \\
    +VL+S & \checkmark & \checkmark & & 55.45 & 71.6 \\
    +VL+S+M & \checkmark & \checkmark & \checkmark & \textbf{55.55} & \textbf{72.3} \\
    \hline
    \end{tabular}
    $}
    \end{center}
\end{table}

The setting in the second row of Table\tab\ref{table:ablation_studies} empirically gives the best performance, and thus we stick to this setting for the following experiments.

\subsubsection{Module Effectiveness}

We ablate the training strategy and the modules in our model to study their effectiveness. 
\textbf{VL} indicates that the RVL-BERT utilizes the external multimodal knowledge learned in the pretext tasks via weight initialization.
\textbf{Spatial} (\textbf{S}) means the spatial module, while \textbf{Mask Att.} (\textbf{M}) stands for the mask attention module.
Table\tab\ref{table:module_effeciveness} shows that each module effectively helps boost the performance. 
The visual-linguistic commonsense knowledge lifts the \textbf{Basic} model by 12\% (or absolute 5\%)  of Recall@50 on VRD dataset, while the spatial module further boosts the model by more than 23\% (or absolute 10\%).
As the effect of the mask attention module is not apparent on the VRD dataset (0.2\% improvement), we also experiment on the SpatialSense dataset (Overall Accuracy) and find the mask attention module provide a relative 1\% boost of accuracy.

\subsection{Quantitative Results on VRD Dataset}

\begin{table}[t!]
    \begin{center}
    \caption{Performance comparison with existing models on VRD dataset. Results of the existing methods are extracted from \cite{lu2016visual} and respective papers.} 
    \label{table:result-vrd}
    \def\arraystretch{1.2}
    \resizebox{0.88\linewidth}{!}{$
    \begin{tabular}{l|cc}
    \hline
    Model & Recall@50 & Recall@100 \\
    \hline
    Visual Phrase \cite{sadeghi2011recognition} & 0.97  & 1.91  \\ 
    Joint CNN \cite{lu2016visual}          & 1.47  & 2.03  \\
    VTransE \cite{zhang2017visual}             & 44.76 & 44.76 \\ 
    PPR-FCN \cite{zhang2017ppr}                & 47.43 & 47.43 \\ 
    Language Priors \cite{lu2016visual}        & 47.87 & 47.87 \\ 
    Zoom-Net \cite{yin2018zoom}                  & 50.69 & 50.69 \\ 
    TFR \cite{jae2018tensorize}                 & 52.30 & 52.30 \\ 
    Weakly (+ Language) \cite{peyre2017weakly} & 52.60 & 52.60 \\ 
    LK Distillation \cite{yu2017visual}          & 55.16 & 55.16 \\ 
    Jung et al. \cite{Jung2018VisualRD} & 55.16 & 55.16 \\
    UVTransE \cite{hung2019union}             & 55.46 & 55.46 \\ 
    MF-URLN \cite{zhan2019exploring}            & 58.20 & 58.20 \\
    HGAT \cite{mi2020hierarchical}              & \textbf{59.54} & \textbf{59.54} \\
    \hline
    \textbf{RVL-BERT}                & 55.55 & 55.55 \\
    \hline
    \end{tabular}
    $}
    \end{center}
\end{table}

\begin{table*}[ht!]
    \begin{center}
    \caption{Classification accuracy comparison on the test split of SpatialSense dataset. Bold font represents the highest accuracy; underline means the second highest. Results of existing methods are extracted from \cite{yang2019spatialsense}. $\dagger$Note that DSRR \cite{9175559} is a concurrent work that was published in August 2020.} 
    \label{table:result-spatialsense}
    \resizebox{1.0\linewidth}{!}{$
    \begin{tabular}{l|c|ccccccccc}
    \hline
    \centering Model & Overall & above & behind & in & in front of & next to & on & to the left of & to the right of & under \\
    \hline
    L-baseline \cite{yang2019spatialsense}   & 60.1 & 60.4 & 62.0 & 54.4 & 55.1 & 56.8 & 63.2 & 51.7 & 54.1 & 70.3 \\  
    PPR-FCN \cite{zhang2017ppr}                & 66.3 & 61.5 & 65.2 & 70.4 & 64.2 & 53.4 & 72.0 & \underline{69.1} & 71.9 & 59.3 \\  
    ViP-CNN \cite{li2017vip}                         & 67.2 & 55.6 & 68.1 & 66.0 & 62.7 & 62.3 & 72.5 & \textbf{69.7} & 73.3 & 66.6 \\  
    Weakly \cite{peyre2017weakly}    & 67.5 & 59.0 & 67.1 & 69.8 & 57.8 & \textbf{65.7} & 75.6 & 56.7 & 69.2 & 66.2 \\  
    S-baseline \cite{yang2019spatialsense}     & 68.8 & 58.0 & 66.9 & \underline{70.7} & 63.1 & 62.0 & 76.0 & 66.3 & \underline{74.7} & 67.9 \\  
    VTransE \cite{zhang2017visual}                   & 69.4 & 61.5 & 69.7 & 67.8 & 64.9 & 57.7 & 76.2 & 64.6 & 68.5 & 76.9 \\  
    L+S-baseline \cite{yang2019spatialsense}  & 71.1 & 61.1 & 67.5 & 69.2 & 66.2 & 64.8 & 77.9 & \textbf{69.7} & \underline{74.7} & \underline{77.2} \\  
    DR-Net \cite{dai2017detecting}                    & 71.3 & \textbf{62.8} & \textbf{72.2} & 69.8 & 66.9 & 59.9 & \underline{79.4} & 63.5 & 66.4 & 75.9 \\  
    DSRR$\dagger$ \cite{9175559} & \textbf{72.7} & 61.5 & \underline{71.3} & 71.3 & 67.8 & \underline{65.1} & \textbf{79.8} & 67.4 & \textbf{75.3} & \textbf{78.6} \\
    \hline
    \textbf{RVL-BERT}                 & \underline{72.3} & \underline{62.5} & 70.3 & \textbf{71.9} & \textbf{70.2} & \underline{65.1} & 78.5 & 68.0 & 74.0 & 75.5 \\
    \hline
    Human Perf. \cite{yang2019spatialsense}    & 94.6 & 90.0 & 96.3 & 95.0 & 95.8 & 94.5 & 95.7 & 88.8 & 93.2 & 94.1 \\  
    \hline
    \end{tabular}
    $}
    \end{center}
\end{table*}

We conduct experiments on VRD dataset to compare our method with existing approaches.
\textbf{Visual Phrase} \cite{sadeghi2011recognition} represents visual relationships as visual phrases and learns appearance vectors for each category for classification. 
\textbf{Joint CNN} \cite{lu2016visual} classifies the objects and predicates using only visual features from bounding boxes.
\textbf{VTransE} \cite{zhang2017visual} projects objects and predicates into a low-dimensional space and models visual relationships as a vector translation. 
\textbf{PPR-FCN} \cite{zhang2017ppr} uses fully convolutional layers to perform relationship detection. 
\textbf{Language Priors} \cite{lu2016visual} utilizes individual detectors for objects and predicates and combines the results for classification. 
\textbf{Zoom-Net} \cite{yin2018zoom} introduces new RoI Pooling cells to perform message passing between local objects and global predicate features. 
\textbf{TFR} \cite{jae2018tensorize} performs a factorization process on the training data and derives relational priors to be used in VRD.
\textbf{Weakly} \cite{peyre2017weakly} adopts a weakly-supervised clustering model to learn relations from image-level labels. 
\textbf{LK Distillation} \cite{yu2017visual} introduced external knowledge with a teacher-student knowledge distillation framework.
\textbf{Jung et al.} \cite{Jung2018VisualRD} propose a new spatial vector with element-wise feature combination to improve the performance.
\textbf{UVTransE} \cite{hung2019union} extends the idea of vector translation in VTransE with the contextual information of the bounding boxes. 
\textbf{MF-URLN} \cite{zhan2019exploring} uses external linguistic knowledge and internal statistics to explore undetermined relationships.
\textbf{HGAT} \cite{mi2020hierarchical} proposes a TransE-based multi-head attention approach performed on a fully-connected graph.

Table\tab\ref{table:result-vrd} shows the performance comparison on the VRD dataset.\footnote{Note that for the results other than \textbf{Visual Phrases} and \textbf{Joint CNN}, Recall@50 is equivalent to Recall@100 (also observed in \cite{lu2016visual,yu2017visual}) because the number of ground truth subject-object pairs is less than 50.}
It can be seen that our \textbf{RVL-BERT} achieves competitive Recall@50/100 (53.07/55.55) compared to most of the existing methods, while lags behind the latest state-of-the-art, such as MF-URLN and HGAT. We note that the use of an additional graph attention network in HGAT and a confidence-weighting module in MF-URLN can be possibly incorporated into our design, while we leave for future work.

\subsection{Quantitative Results on SpatialSense Dataset}

We compare our model with various recent methods, including some methods that have been compared in the VRD experiments.
Note that \textbf{L-baseline}, \textbf{S-baseline} and \textbf{L+S-baseline} are baselines in \cite{yang2019spatialsense} taking in simple language and/or spatial features and classifying with fully-connected layers. 
\textbf{ViP-CNN} \cite{li2017vip} utilizes a phrase-guided message passing structure to model relationship triplets. 
\textbf{DR-Net} \cite{dai2017detecting} exploits statistical dependency between object classes and predicates. 
\textbf{DSRR} \cite{9175559} is a concurrent work\footnote{Originally published in August 2020.} that exploits depth information for relationship detection with an additional depth estimation model.
The \textbf{Human Performance} result is extracted from \cite{yang2019spatialsense} for reference. 

Table\tab\ref{table:result-spatialsense} shows that our full model outperforms almost all existing approaches in terms of the overall accuracy and obtains the highest (like {\fontfamily{qcr}\selectfont in} and {\fontfamily{qcr}\selectfont in front of}) or second-highest accuracy for several relationships. While the concurrent work DSRR achieves a slightly higher overall recall, we expect our model to gain another performance boost with the additional depth information introduced in their work. 

\subsection{Qualitative Results of Visual-Linguistic Commonsense Knowledge}

\begin{figure*}[ht!]
\centering
\includegraphics[width=0.9\textwidth]{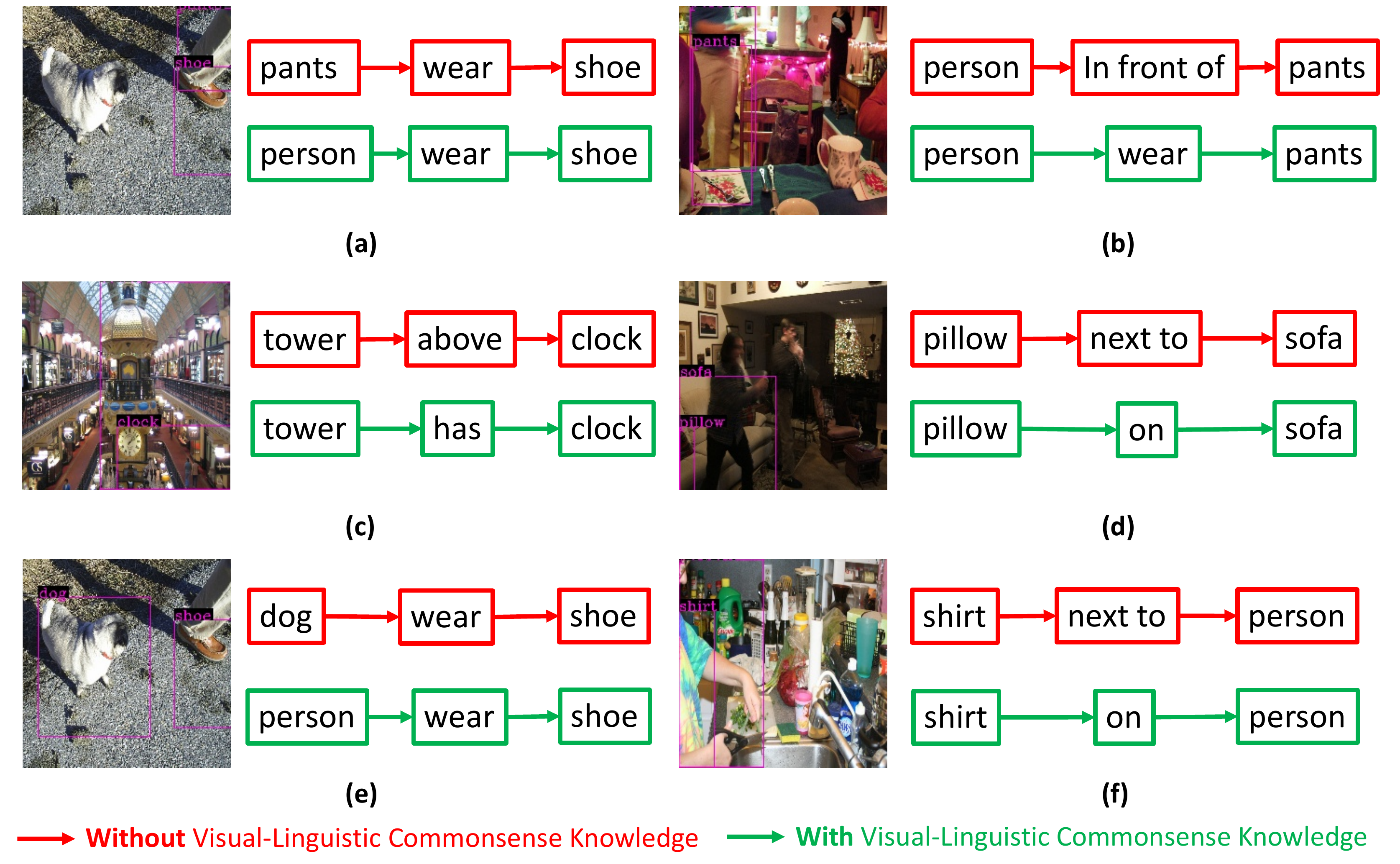}
\caption{Qualitative comparisons between predicting visual relationship with or without visual-linguistic commonsense knowledge. Red boxes and arrows denotes predicting with the model without the knowledge, while the green boxes and arrows means predicting with the knowledge. This visualizations are performed during testing on the VRD dataset \cite{lu2016visual}.}
\label{fig:vlk-visualization}
\end{figure*}

Figure\fig\ref{fig:vlk-visualization} shows qualitative comparisons between predicting visual relationships with and without the visual-linguistic commonsense knowledge in our model. 
Especially, the example (a) in the figure shows that, with \textit{linguistic} commonsense knowledge, a {\fontfamily{qcr}\selectfont person} is more likely to {\fontfamily{qcr}\selectfont wear} a {\fontfamily{qcr}\selectfont shoe}, rather than {\fontfamily{qcr}\selectfont pants} to {\fontfamily{qcr}\selectfont wear} a {\fontfamily{qcr}\selectfont shoe}.
That is, the conditional probability $p(\text{\fontfamily{qcr}\selectfont wear}|(\text{\fontfamily{qcr}\selectfont person-shoe}))$ becomes higher, while $p(\text{\fontfamily{qcr}\selectfont wear}|(\text{\fontfamily{qcr}\selectfont pants-shoe}))$ becomes lower and does not show up in top 100 confident triplets after observing the linguistic fact.
Same applies to the example (b) (where {\fontfamily{qcr}\selectfont person-wear-pants} is more appropriate than {\fontfamily{qcr}\selectfont person-in front of-pants}) and the example (c) (where  {\fontfamily{qcr}\selectfont tower-has-clock} is semantically better than {\fontfamily{qcr}\selectfont tower-above-clock}). 
On the other hand, as the {\fontfamily{qcr}\selectfont person} in the example (e) is visually occluded, the model without \textit{visual} commonsense knowledge prefers to {\fontfamily{qcr}\selectfont dog-wear-shoe} rather than {\fontfamily{qcr}\selectfont person-wear-shoe}; however, our model with the visual knowledge knows that the occluded part is likely to be a person and is able to make correct predictions. 
Same applies to the example (d) (where both {\fontfamily{qcr}\selectfont pillow} and {\fontfamily{qcr}\selectfont sofa} are not clear) and (f) (where {\fontfamily{qcr}\selectfont person} is obscure). 
These examples demonstrate the effectiveness of our training strategy of exploiting rich visual and linguistic commonsense knowledge by pre-training on unlabeled visual-linguistic datasets.

\subsection{Qualitative Results of Mask Attention Module}

\begin{figure*}[t!]
\centering
\includegraphics[width=0.9\textwidth]{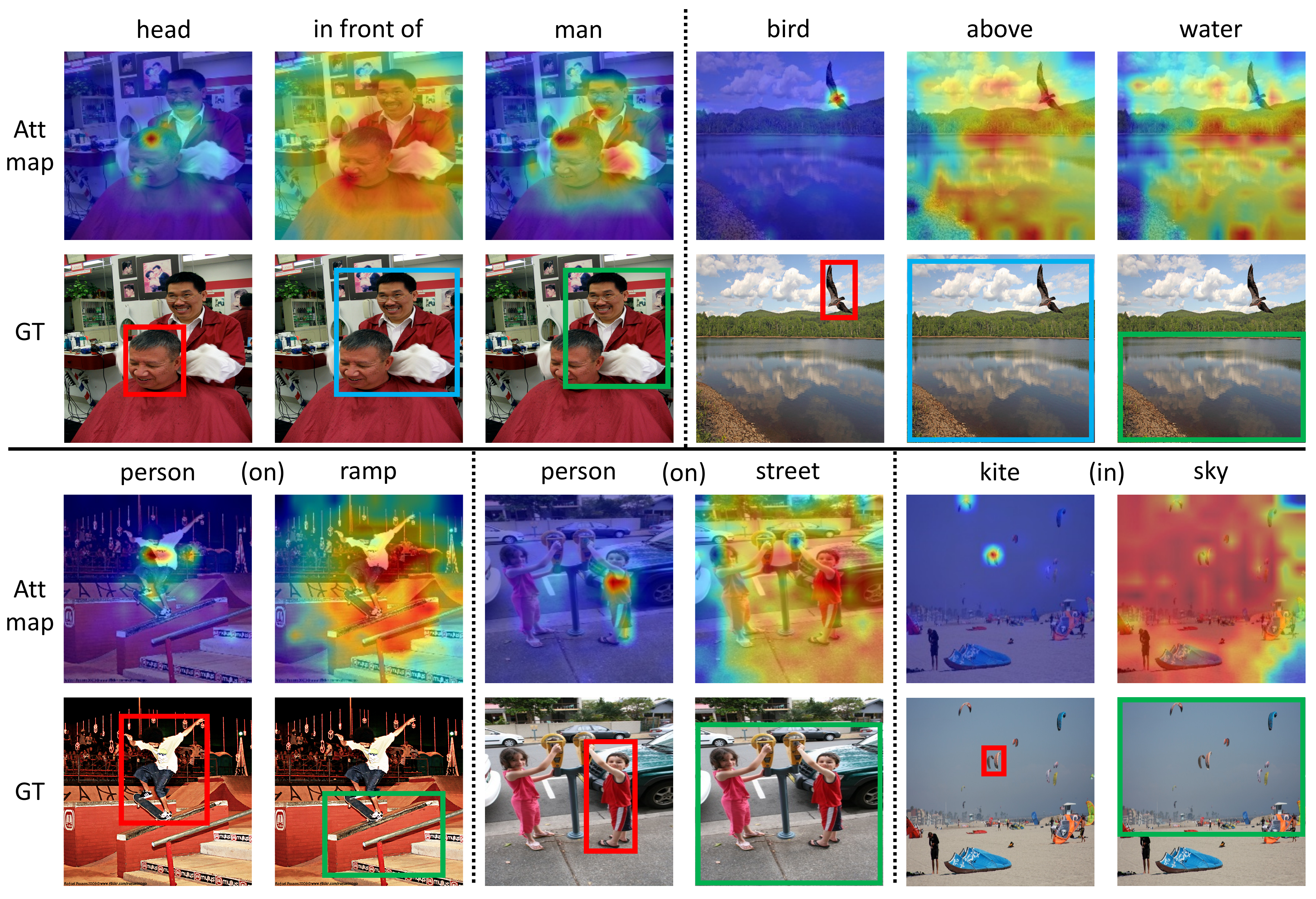}
\caption{Attention map visualization of SpatialSense (the first two rows) and VRD dataset (the last two rows). For each example, the first row shows predicted attention maps while the second shows ground truth bounding boxes.}
\label{fig:heatmap-vis-full}
\end{figure*}

The mask attention module aims to teach the model to learn and predict the attention maps emphasizing the locations of the given object labels. 
To study its effectiveness, we visualize the attention maps that are generated by the mask attention module during testing on the both datasets in Figure\fig\ref{fig:heatmap-vis-full}.
The first two rows show two examples from SpatialSense, while the last two rows show three examples from the VRD dataset. 
Since the input embeddings of the model for the SpatialSense dataset in the form of triplets {\fontfamily{qcr}\selectfont subject-predicate-object}, and the VRD dataset in the form of doublets {\fontfamily{qcr}\selectfont subject-object} are different, three and two attention maps are generated for each example, respectively.

For both datasets, the model is actively attending to the region that contains the target object.
Especially for the triplet data from SpatialSense, the model is also looking at the union bounding boxes which include cover both subjects and objects. 
For example, for the top-left example {\fontfamily{qcr}\selectfont head-in front of-man} in Figure \ref{fig:heatmap-vis-full}, mask attention first looks at the person's {\fontfamily{qcr}\selectfont head} who is getting a haircut, followed by attending to the joint region of {\fontfamily{qcr}\selectfont head} and the barber ({\fontfamily{qcr}\selectfont man}), then finally focus on the barber {\fontfamily{qcr}\selectfont man}.
We also observe that, for objects (classes) with larger size, mask attention tends to look more widely at the whole image.
For instance, {\fontfamily{qcr}\selectfont water} of the top-right example and {\fontfamily{qcr}\selectfont sky} of the bottom-right example attend to almost the whole image. We conjecture that this is due to the larger size of objects or regions making it harder to learn to focus on the specific target areas. 
In addition, We also find the mask attention module learns better with triplet inputs than doublets inputs and this is assumedly because the additional examples of union boxes provide more contexts and facilitate the learning process.

\section{Conclusions}

In this paper, we proposed a novel visual relationship detection system named RVL-BERT, which exploits visual commonsense knowledge in addition to linguistic knowledge learned during self-supervised pre-training.
A novel mask attention module is designed to help the model learn to capture the distinct spatial information and a spatial module is utilized to emphasize the bounding box coordinates. 
Our model is flexible in the sense that it can be solely used for predicate classification or cascaded with any state-of-the-art object detector.
We have shown that the effectiveness of the proposed modules and its competitive performance through ablation studies, quantitative and qualitative experiments on two challenging visual relationship detection datasets.

For future work, we propose that the underlying model (\emph{i.e.}, BERT) can be re-designed to accommodate all proposals so that it could forward all pairs altogether instead of pairwise predictions.
We also anticipate that more visual and/or linguistic commonsense databases could be utilized for pre-training our model.

\begin{appendix}
\section*{Appendix}
\subsection{Results on Visual Genome}
We provide additional experimental results on Visual Genome (VG) dataset \cite{krishna2017visual}. We follow \cite{xu2017scene,zellers2018neural,tang2020unbiased} to adopt the most widely-used dataset split which consists of 108K images and includes the most frequent 150 object classes and 50 predicates. 
When evaluating visual relationship detection/scene graph generation on VG, there are three common evaluation modes including (1) Predicate Classification (PredCls): ground truth bounding boxes and object labels are given, (2) Scene Graph Classification (SGCls): only ground truth boxes given, and (3) Scene Graph Detection (SGDet): nothing other than input images is given. We experiment with PredCls, which is a similar setting to what we perform on VRD dataset \cite{lu2016visual} and SpatialSense \cite{yang2019spatialsense} dataset. 

\begin{table}[h!]
    \begin{center}
    \caption{Performance comparison with existing models on VG dataset. Results of the existing methods are extracted from the respective papers. Models are evaluated in PredCls mode.}
    \label{table:result-vg}
    \def\arraystretch{1.2}
    \resizebox{0.88\linewidth}{!}{$
    \begin{tabular}{l|cc}
    \hline
    Model & Recall@50 & Recall@100 \\
    \hline
    Joint CNN \cite{zhang2017visual}            & 27.9 & 35.0 \\
    Graph R-CNN \cite{yang2018graph}            & 54.2 & 59.1 \\
    Message Passing \cite{xu2017scene}          & 59.3 & 61.3 \\
    Freq \cite{zellers2018neural}               & 59.9 & 64.1 \\
    Freq+Overlap \cite{zellers2018neural}       & 60.6 & 62.2 \\
    SMN \cite{zellers2018neural}                & 65.2 & 67.1 \\
    KERN \cite{chen2019knowledge}               & 65.8 & 67.7 \\
    VCTree \cite{tang2019learning}              & \textbf{66.4} & \textbf{68.1} \\
    \hline
    \textbf{RVL-BERT}              & 62.9 & 66.6 \\
    \hline
    \end{tabular}
    $}
    \end{center}
\end{table}

Comparison results on VG are presented in Table \ref{table:result-vg}, where our proposed RVL-BERT achieves competitive results.
Note that as mentioned in section \ref{sec:datasets}, visual relationship detection can be biased as predicates could be "guessed" accurately given explicit correlations between object labels and predicates. 
Both SMN \cite{zellers2018neural} and KERN \cite{chen2019knowledge} exploit this property and use the frequency bias and object co-occurrence, respectively. 
However, the usage of bias could reversely undermine the capability of generalization which has been demonstrated by comparing mean recall in recent works (e.g., \cite{tang2020unbiased}).
\end{appendix}

\bibliographystyle{IEEEtran}
\bibliography{rvl-bert}

\begin{IEEEbiography}
[{\includegraphics[width=1in,height=1.25in,clip,keepaspectratio]{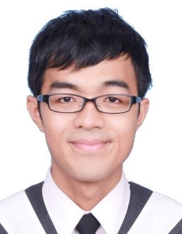}}]{Meng-Jiun Chiou} received his B.Sc.~degree in electrical and computer engineering from National Chiao Tung University, Hsinchu, Taiwan, in 2016. He is currently a computer science PhD candidate at the National University of Singapore and is under the supervision of both Prof. Roger Zimmermann at the Department of Computer Science and Prof.~Jiashi Feng at the Department of Electrical and Computer Engineering. His research is focused in computer vision and machine learning, especially, scene understanding with visual relationship detection. He also has experience in human-object interaction detection, visual relationship guided image captioning, zero-shot indoor localization and weakly-labeled audio tagging. He has published in top multimedia conferences including ACM MM (’19 \& ’20).
\end{IEEEbiography}

\begin{IEEEbiography}
[{\includegraphics[width=1in,height=1.25in,clip,keepaspectratio]{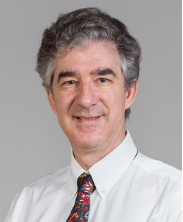}}]{Roger Zimmermann} (M’93–SM’07) received his M.S.~and Ph.D.~degrees from the University of Southern California (USC) in 1994 and 1998, respectively. He is a professor with the Department of Computer Science at the National University of Singapore (NUS). He is also a Deputy Director with the Smart Systems Institute (SSI) and a key investigator with the Grab-NUS AI Lab. He has coauthored a book, seven patents, and more than 300 conference publications, journal articles, and book chapters. His research interests include streaming media architectures, multimedia networking, applications of machine/deep learning, and spatial data analytics. He is an associate editor for IEEE MultiMedia, ACM Transactions on Multimedia Computing, Communications, and Applications (TOMM), Springer Multimedia Tools and Applications (MTAP), and IEEE Open Journal of the Communications Society (OJ-COMS). He is a distinguished member of the ACM and a senior member of the IEEE.
\end{IEEEbiography}

\begin{IEEEbiography}
[{\includegraphics[width=1in,height=1.25in,clip,keepaspectratio]{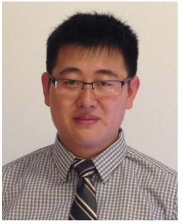}}]{Jiashi Feng} (Member, IEEE) received the B.Eng. degree from the University of Science and Technology, China, in 2007, and the Ph.D. degree from the National University of Singapore in 2014. He was a Postdoctoral Researcher with the University of California from 2014 to 2015. He is currently an Assistant Professor with the Department of Electrical and Computer Engineering with the National University of Singapore. His current research interests include machine learning and computer vision techniques for large-scale data analysis. Specifically, he has done work in object recognition, deep learning, machine learning, high-dimensional statistics, and big data analysis.
\end{IEEEbiography}

\EOD

\end{document}